# A K-MEANS, WARD AND DBSCAN REPEATABILITY STUDY

**Anthony Bertrand, Engelbert Mephu Nguifo, Violaine Antoine and David R.C. Hill**

Université Clermont Auvergne, Clermont Auvergne INP, ENSM St Etienne, CNRS, LIMOS, F-63000 Clermont–Ferrand, France

anthony.bertrand@uca.fr, engelbert.mephu_nguifo@uca.fr, violaine.antoine@uca.fr, david.hill@uca.fr [1]

**Abstract :**

*Reproducibility is essential in machine learning because it ensures that a model or experiment yields the same scientific conclusion. For specific algorithms repeatability with bitwise identical results is also a key for scientific integrity because it allows debugging. We decomposed several very popular clustering algorithms: K-Means, DBSCAN and Ward into their fundamental steps, and we identify the conditions required to achieve repeatability at each stage. We use an implementation example with the Python library scikit-learn to examine the repeatable aspects of each method. Our results reveal inconsistent results with K-Means when the number of OpenMP threads exceeds two. This work aims to raise awareness of this issue among both users and developers, encouraging further investigation and potential fixes.*

## 1 Introduction

Experimental science relies on the reproducibility of experiments. When the results are reproducible, they are analyzed and processed to draw scientific conclusions. It is therefore important to ensure that our results are similar from one experiment to the next. Reproducibility issues introduced by the Computer Science stack of tools (hardware, software, libraries and execution environments) and procedures are now affecting all scientific disciplines [1]. When run under the same conditions, machine learning and Artificial Intelligence models or experiments should give the same scientific conclusions. This requirement is critical for several reasons: scientific integrity, model validation and comparison but also for debugging and software development. In the latter, we also need bitwise identical traces and results. Indeed, developers need deterministic behavior to identify implementation bugs, optimize performance, and ensure consistent behavior across platforms and hardware configurations.

---

[1] Corresponding author

However, with modern machine learning frameworks which hide technical details, it becomes more and more difficult to control the pseudorandom sources hidden in the machine learning stochastic models [2]. Parallelization, aimed at accelerating calculations introduces additional challenges for reproducibility [3]. To overcome repeatability issues, some researchers perform multiple replications of the same experiment without realizing the bias they introduce when they do not master the proper management of parallel random sources. However, this flawed approach is common, and many scientists present just biased statistics on their results. Before working on reproducibility, we often need to have repeatable numerical experiments. Bitwise identical results from run to run on the same machine and environment with the same code input data is mandatory for debugging. This corresponds to the ACM definition of repeatability with a stated precision equal to 0 (identical results). With modern machine learning, we often must fight to keep this precious repeatability.

In this paper, we focus on clustering methods. They are data analysis techniques that group data, based on shared characteristics. They fall under unsupervised learning, where the machine independently identifies patterns without predefined labels. According to [4], we can classify clustering methods in two categories: Hierarchical and Partitional. Hierarchical clustering methods build cluster iteratively with two approaches. The first approach called Agglomerative clustering uses bottom-up approach, starting with every data as its own cluster, and merging similar clusters iteratively until only one cluster remains. The second approach, called divisive clustering, uses a top-down approach, starting with one cluster and dividing it into multiple ones until each object forms a cluster. These two approaches end up building dendrograms. Partitional clustering optimizes objective functions. Finally, we retain three methods: (1) Distance-based clustering - which often uses the Euclidean distance between data points ; (2) Model-based clustering – which tries to optimize a mathematical model ; (3) Density-based clustering - which uses density function to find clusters.

One of the most used Python library for machine learning is scikit-learn [5]. It is an open-source machine learning library. It offers many useful tools for AI developers, such as generation of synthetic data, supervised and unsupervised learning algorithms and metrics to evaluate created models. Due to its versatility and ease of use, this library is adopted across many scientific domains.

In this paper, we investigate repeatability issues with three well-known algorithms from three different clustering methods: K-Means for distance-based clustering, DBSCAN for density-based clustering, and the Ward method for agglomerative clustering. These algorithms are very popular and cover different categories of the taxonomy above. We organized our paper as follow: section 2 gives definitions of reproducibility, repeatability, and their importance in experimental sciences. Section 3 presents pitfalls ML scientists must avoid to keep their work repeatable. Section 4 describes the three chosen algorithms. Section 5 presents our experiments and results we obtained. Section 6 discusses these results and Section 7 concludes the article.

## 2 Reproducibility and repeatability

In our article, we focus on both reproducibility and repeatability of results. To fully understand these issues, it is necessary to address the question of reproducibility in science.

Reproducibility is one of the pillars of Science. For philosophers of science, it is even one of the criteria that distinguishes Science from pseudoscience. Here is the definition for reproducibility established by the ACM in 2020: "The measurement can be obtained with stated precision by a different team using the same measurement procedure, the same measuring system, under the same operating conditions, in the same or a different location on multiple trials. For computational experiments, this means that an independent group can obtain the same result using the author's own artifacts." ([ACM Badges 2020](#)).

The term "reproducibility crisis" emerged and gained popularity in the past decade. This crisis is due to an excessive number of articles whose experiments and results are not reproducible, and quite often reviewers are unable to reproduce experiments. The sources of non-reproducibility are numerous and are not explored in our article. This crisis has led to rapid developments in the field of reproducibility, and several surveys report developments in this area [6]. In the field of machine learning, many studies demonstrate this problem, particularly [7]. They made the hypothesis that AI work is not enough documented to be reproducible. To evaluate this, they set variables to describe how much documented a scientific work is. Then they surveyed 400 research papers published in top AI conferences and used their metrics. None of the papers was fully reproducible. [8] is a paper dealing with AI for breast cancer screening. It denounces the aspects that must be clarified by article authors to be able to reproduce their work. This article is a good educational example for reproducibility. Other works noticed the confusion in the terminology used in reproducible research, and they tried to introduce frameworks to facilitate validation [9,10].

Repeatability is different from reproducibility, but it is an important aspect, particularly when we deal with computer code. Here is the definition for repeatability proposed by the ACM: "The measurement can be obtained with stated precision by the same team using the same measurement procedure, the same measuring system, under the same operating conditions, in the same location on multiple trials. For computational experiments, this means that a researcher can reliably repeat her own computation." (ACM Badges 2020). For the setting up of programs used to produce Science, we need to be able to debug, and this implies that the stated precision is equal to zero. Indeed, we need bitwise identical results from run to run in the same conditions, otherwise how do you debug? The computers we use are deterministic machines, and after the first years of hardware and software development, it became easy to obtain the same results every time with a deterministic stack of tools. This was taken for granted for many decades. But since the development of sophisticated frameworks to hide complexity, hardware optimization in microprocessors, the massive usage of hyperthreading and the trickiness of parallel random number generation, many scientists do not even notice that they are producing different results from run to run. The software stack they use drives them into biased and non-repeatable experiments.

Reproducibility is less demanding since you just expect "similar" results leading to the same scientific conclusion, but it also relies on a sound repeatable program. Ideally, all experiments should be reproducible. However, many reproducibility issues in machine learning are largely due to repeatability issues, meaning that the code is developed on quicksand. [11] tried to obtain a repeatable work in deep learning with a GPU. They used Pytorch and Tensorflow/Keras as ML frameworks. They show their inability to obtain repeatable results with Tensorflow 2.1/Keras 2.3 in an image segmentation task, even when setting properly the seeds of the random number generators and disabling CUDA flags responsible for non-repeatable results. [12] worked on repeatability in deep reinforcement learning. They first identified the sources of non-determinism to control to obtain repeatable results. Then, they studied the effect of these sources of non-determinism.

In our paper, we focus on repeatability for three main families of clustering algorithms. When conducing experiments to make repeatable work with clustering methods to propose valuable scientific contributions, we searched for the existence of repeatable traps to warn users. This is where we found an interesting issue mentioned in the scikit-learn's GitHub website (https://github.com/scikit-learn/scikit-learn/issues/27518). This issue describes inconsistent results when using parallelism with OpenMP for the K-Means algorithm. We found it still current in the 1.7.0 version of scikit-learn. OpenMP is an application programming interface (API) for shared-memory multiprocessing programming. When correctly used, it is known to be deterministic. To our knowledge, K-Means is the only scikit-learn algorithm with a repeatability problem. We emit the hypothesis that the scikit-learn implementation of K-Means contains parallelism mismanagements with OpenMP.

# 3 Repeatability pitfalls

In this section, we explain the possible responsible of hidden randomization in computation. A person willing to obtain bitwise repeatability must be aware of these issues.

## 3.1 Random sources

When working with stochastic experiments, using randomness in calculations, it is necessary to master random sources for repeatable experiments on computers. This is the precise purpose of pseudorandom number generators, whose behavior is designed to be deterministic models of randomness.

In this domain, we must be careful with the widely used grandfather terminology of seeds and seeding. They are still in use in major APIs (Application Programmer Interfaces), but they can be really confusing for modern generators. Indeed, sound and modern generators cannot be reliably initialized by a "poor integer". The initial statuses of modern generators are much bigger and or structured. There is no bijection between integers and the set of initial statuses of modern generators [2]. In addition, the API and the status structure might change from one language/library to the other even when considering the same generator and we may find different traces with the same generator and the same seed! To guarantee a good initialization we must use the complete state of the generator. Machine learning APIs often oversimplifies the seeding. If we go back to the generator implementations, they provide a way to get/set the precise state of a generator, the latter taking the shape of a more complex object. It can be for instance a list of integers for the Mersenne Twister generator [13], or a couple of key/counter for the Philox generator [14]. Even if something changes in the library implementation, or if we want to use another library, the state guarantees the same generator.

## 3.2 Parallelism

Modern computers have CPUs with multiple cores. This allows them to run different tasks in parallel to increase program's speed/efficiency. To take advantage of this, Python often uses compiled files written in C or C++ with the OpenMP library. The code written with these two languages is compiled and therefore runs faster than pure Python code. The OpenMP library consists of a set of compiled directives influencing the program behavior. It provides for instances ways to optimize loops, using multiple threads. Programmers must implement these optimizations correctly to avoid thread synchronization problems and race conditions.

One common mistake to avoid for repeatable results is to keep the order of floating-point operations. They are non-associative and lead to inconsistent results, as illustrated by Figure 1.

```
>>> 0.1+(0.2-0.1)
0.2
>>> (0.1+0.2)-0.1
0.20000000000000004
```

Figure 1: Python Example of non-associativity in floating-point operations

One easy solution when we notice inconsistent results with a compiled code is for instance to disable parallelism induced by OpenMP. We can set the maximum number of threads usable by OpenMP in system variables to 1 or use context managers in Python with threadpoolctl library for example. However, it is possible to achieve repeatable multithreading, as demonstrated [15] with OpenMP, by defining the behavior using flags during compilation.

## 4 Clustering methods

In this section, we give a formal definition of the studied algorithms. Then we describe them with simple step to follow and explain how repeatability pitfalls defined in Section 3 can intervene in the algorithm.

### 4.1 Distance-based clustering: study of K-Means

The K-Means algorithm has been proposed by several researchers as summarized by [16], but the name generally refers to the Lloyd algorithm [17] which is the most used algorithm. It has some flaws, like its linearity, creating spherical clusters due to the use of Euclidean metrics as a distance measure. This is a greedy algorithm, meaning that for each step of the algorithm, it will make the locally optimal choice. It then converges to a local minimum, and needs multiple runs with different initializations to obtain a better result.

However, despite these flaws, it is widely used in data mining for its simplicity and low polynomial complexity. K-Means solves the partitioning problem by minimizing the sum of square error (SSE) of each cluster. Let $X = (x_i) \in \mathbb{R}^{d*n}$ our $n$ data points with $d$ features, $C = (c_k) \in \mathbb{R}^{d*K}$ our K centers and $p_{ik}$ a Boolean representing the membership of point $x_i$ to cluster represented by the center $c_k$. The SSE, also known as minimum sum-of-squares clustering (MSSC) [18] is:

$$\min_{p,c} \sum_{i=1}^{n} \sum_{k=1}^{K} p_{ik} \|x_i - c_k\|^2$$

$$\text{subject to} \begin{cases} 1. \sum_{k=1}^{K} p_{ik} = 1, \forall i \in [1 \dots n] \\ 2. p_{ik} \in \{0,1\} \\ 3. \sum_{i=1}^{n} p_{ik} \geq 1, \forall k \in [1 \dots K] \end{cases}$$

The first constraint matches each point with a center. The third constraint ensures a clustering with K clusters. This algorithm can be separated into three steps:

1. **Initialization**: We initialize the centers $c_k$.

2. **Assignment**: For each point $x_i$, we calculate the Euclidean distances and assign it to the cluster represented by the closest center $c_k$.

3. **Update**: We replace each center $c_k$ by the average of the points which compose the cluster

Steps 2 and 3 are repeated until convergence or stopping of the algorithm. Note that K-Means is a heuristic to the MSCC objective function that omits the third constraint. This omission allows empty clusters, leading to degenerated results [19]. Let us see the repeatability aspect of each step.

**Initialization**: The algorithm is very sensitive to initialization. Finding the minimum of the objective function in NP-hard. K-Means is a heuristic approach. This is why researchers generally execute the algorithm several times with different initializations to keep the most efficient execution. During this step, we initialize cluster centers randomly. This needs correct management of the random source. It must be set with a known initial status to be repeated.

**Assignment**: When calculating distances, it is important to define in advance the behavior of the algorithm in case of equidistance of a point with several centroids. We must ensure a deterministic selection (example: assign the point to the centroid that occupies the smallest index in the list of centroids).

**Update**: We need to compute the average of all the points belonging to the cluster. The centroid of each cluster will then update their position accordingly. During this step, there could be an empty cluster due to a centroid being too far away from the data. The position of the centroid of this cluster will not be able to be updated. The solution is then considered degenerated. In order to have the correct number of clusters, rules must be set for this specific case scenario.

### 4.2 Density-based clustering: study of DBSCAN

The DBSCAN algorithm (density-based spatial clustering of applications with noise) proposed by [20] is an algorithm that uses density to group or separate data. This allows to discover clusters with various shapes, offering non-linear clustering possibilities, unlike K-Means, for example. There is no need to specify the number of clusters, and it has the capacity to handle outliers. However, the default algorithm struggles with data exhibiting highly variable neighborhood densities, and data with overlapped clusters.

Let $P \subset powerset(\chi)$ be the space of solutions such that $C = \{C_1, \dots, C_l\}$ partitions the dataset. Let $d_{db}^{\mu}(p, q)$ be the DBSCAN-distance, which gives the smallest $\varepsilon$, s.t. two points are in the same DBSCAN cluster. Then we define the $\varepsilon$-density-based-clustering ($\varepsilon$DBC) objective as [21]:

$$\min_{\substack{C \subset P \\ d_{db}^{\mu}(p,q) \leq \varepsilon \forall p,q \in C_i \forall C_i \in C}} |C|$$

The algorithm requires two input parameters: the distance $\varepsilon$ defining the neighborhood of a point, and a number of points *MinPts* defining the number of points that must be part of this neighborhood to define the point as a core point. The algorithm classifies the data into three categories:

- Core points, whose neighborhood is dense (i.e., the number of points present within a radius $\varepsilon$ is greater than or equal to *MinPts*).

- Border points, which belong to the neighborhood of a core point without being one itself (i.e., the number of points present within a radius $\varepsilon$ is less than *MinPts*).

- Noise points, which are neither core nor border points.

The algorithm can be separated into 3 steps [22]:

1. **Identification**: For each point, calculate the neighborhood and identify the core points.

2. **Creation**: Among the core points, group those which are neighbors into clusters.

3. **Assignment**: Among the non-core points, if a point is close to a core point, it becomes a border point and is assigned to the cluster. Otherwise, it becomes a noise point.

**Identification**: We need to check for each point if it is a core point. To perform this, we compute how many neighbors each point has. A point has a neighbor if its distance from another point is less than $\varepsilon$. If a point has *MinPts* neighbors, we consider it as a core point. We should not have repeatability issues except if we have troubles with floating point operations (linked to dynamic execution inside microprocessors for optimization purposes). We can think of a rare case scenario where a point has a distance of exactly $\varepsilon$ with another point. If we consider what we said in the parallelism section (section 3.2), we could have the result of the distance equal to $\varepsilon$ + *error*. The point would sometimes be considered as a neighbor, and sometimes not.

**Creation**: If a core point belongs to the neighborhood of another core point, we put them in the same cluster. Same as before, there are no repeatable issues except if we have indeterminism during floating point operations.

**Assignment**: the neighbor's exploration order is very important. If a border point *p* has two neighbors, *a* and *b*, which are core points belonging to different clusters, border point *p* can be added to the cluster of either point *a* or point *b*. This order must be set in advance to obtain the same result each time the algorithm is run.

### 4.3 Agglomerative clustering: study of Ward

Agglomerative clustering methods create nested clusters in a tree-like structure with a bottom-up approach. This allows clusters to be formed without specifying their number, by grouping the closest clusters together using a distance function. The Ward method [23] is a famous hierarchical clustering method. Just like K-Means, it seeks to minimize the variance when merging two clusters. Let $q_i, i = [1 \dots K]$ be our cluster at step n-K, with $q_i^*$ the center of the cluster. Let $q_i, i = [1 \dots K], j = [1 \dots K], i \neq j$ the fusion of cluster *i* and *j*, with $q_{ij}^*$ the center of this cluster. At each step, we try to minimize:

$$\min_{\substack{i,j \\ i \neq j}} \frac{1}{|q_{ij}|} \sum_{x \in q_{ij}} (x - q_{ij}^*)^2$$

The greedy nature of the algorithm makes the choice at each next step the optimal solution, but may result in non-optimal clustering. Here are the main steps:

1. **Initialization**: Each point is its own cluster.

2. **Update**: Compute the predicted variance for each pair of clusters equal to the variance we would have if we merge the clusters.

3. **Merge**: Merge the two most similar clusters (i.e. clusters that gives the minimum variance).

Steps 2 and 3 continue until only one cluster remains. The grouping can then be represented using a dendrogram.

Here, the objective function is the mean squared error. At each step, we choose the two clusters that give the smallest mean squared error (MSE).

**Initialization**: This step is repeatable. Compared to the other algorithms, there is no choice to make here. Every point becomes a cluster at initialization step.

**Update**: We compute the predicted variance for each pair of clusters. Only the non-associativity of floating-point operation can cause troubles.

**Merge**: During the merging step, it is necessary to define the behavior of the algorithm when several pairs of clusters have the same distance. For example, choose the clusters with the smallest indices.

### 4.4 Summing up

Here is an array summing up the important steps and the repeatable aspects that need attention.

We can split repeatable aspects into two categories:

- Random number generator problems (RNG), which include the use of random numbers during the execution. This problem occurs when shuffling data or using a random initialization.
- Operation order problems designate floating-point operations problems that lead to the loss of bitwise repeatability.

Table 1: Recap of repeatable issues that may appear in each algorithm steps

| Algorithms \ Aspects | RNG | Operation order |
|---|---|---|
| **K-Means** | Initialization | Assignment, Update |
| **DBSCAN** |  | Identification, Creation, Assignment |
| **WARD** |  | Update, Merge |

Here, K-Means is the only algorithm using random numbers during initialization process. Most of the problems come from the operation order because of our IEEE754 standard for floating-point operations. These problems exist because of the limitation of our machine. We must think about them when developing our programs.

## 5 Experiments

### 5.1 Hardware and software

For hardware, we use two CPU Xeon Platinum 8470 (104 workers, 208 logical threads).

For software, we use Python 3.11.2, scikit-learn 1.7.0, NumPy 2.3.1, OpenMP 4.5 and OpenBLAS 0.3.29. For more information about libraries version, head on to our Gitlab repository: https://gitlab.limos.fr/anbertrand1/repeatability_quest_clustering.

### 5.2 Method

*5.2.1 Energy measurement*

To measure the energy consumption, we designed a minimalist Python class to retrieve RAPL values during each replication loops. Python libraries like pyRAPL or pyjoules have a similar behavior. However, they do not take into account the reset of RAPL counters. Because of this, we can obtain negative results with these libraries. That is what motivated us to develop a simple solution adapted to our configuration. Our method reads RAPL counters at the start and end of experiment. It then computes energy consumption by subtracting the end value and the start value. If the result is negative, we add the maximum value a counter is supposed to handle. This method allows us to handle negative values when a reset happens during the experiment. If multiple resets happen, it won't work. We ensure the mean duration of one experiment is less than the time it takes for a counter to reset. We consider our method to be good enough, but we plan to improve it in the future.

We also use the perf tool, to measure the energy consumption of one script. It gives us the energy consumption of all algorithms on one specific dataset. It is mainly used for verification with our values.

*5.2.2 Datasets*

We chose five datasets from the UC Irvine Machine Learning Repository (https://archive.ics.uci.edu). Clustering accuracy do not matter, our purpose being to show traces of non-repeatability. We think it is still a good thing to have labeled data people often use to work with. We completed the five datasets with a generated one thanks to the scikit-learn library. This last dataset allows us to have custom data with a defined number of instances, features and clusters. Table 2 shows the characteristics of the chosen datasets.

Table 2: Sum-up of the chosen datasets

| Parameters / Datasets | # Instances | # Features | # Classes |
|---|---|---|---|
| Iris | 150 | 4 | 3 |
| Wine | 178 | 13 | 2 |
| Breast cancer | 569 | 30 | 2 |
| Letter recognition | 20 000 | 16 | 26 |
| Default of credit card clients | 30 000 | 23 | 2 |
| Blob dataset (generated) | 60 000 | 2 | 10 |

For the generated dataset, we used the make_blobs function of scikit-learn with these parameters:
- n_samples       = 60 000,
- n_features      = 2,
- centers         = 10,
- cluster_std     = 0.7,
- random_state    = 42.

For each dataset, we applied a Min-Max Scalar to scale features between 0 and 1.

*5.2.3 Execution*

For each dataset, we run the three algorithms described in this article with scikit-learn library: K-Means, DBSCAN and Ward. To investigate the repeatable problems of Sklearn's K-Means, we added Scipy's K-Means. We focus on multiple things:
- Repeatability of the results.

- Duration of the execution depending on the number of OpenMP threads used.
- Energy consumption of the execution depending on the number of OpenMP threads used.

To do so, we run 30 times each combination of dataset, algorithm and number of OpenMP threads. We chose to measure the time and energy cost of the 30 executions because some methods are too fast to be accurately measured and because we have then enough experiments according to the Central Limit theorem if we need precise statistics.

Here is a sum-up of the execution.

- *Algorithms*: sklearn_dbscan, sklearn_kmeans, sklearn_ward, scipy_kmeans
- *Datasets*: iris, wine, breast_cancer, letter_recognition, credit_card, generated
- *OpenMP threads*: [1, 4, 16, 64, 128, 192]
- Replication: 30

For hyperparameters (HP) selection, we chose only parameters that work without trying to obtain better results.

- For K-Means
    - n_clusters = n_classes
    - rand_init = 42
    - n_init = 5
- For Ward
    - n_clusters = n_classes
- For DBSCAN, HP selection is crucial to obtain a valid clustering. Bad HP selection can lead to a result where every point is considered as outliers. To avoid this, we used a method described in [24]. We chose min_samples=2*n_features and find eps empirically by following [24]. Here are the {eps, min_samples} considered for each dataset:

Table 3: Hyperparameters for DBSCAN and for each dataset

| Name of the datasets | Epsilon | Min_samples |
|---|---|---|
| Iris | 0.2 | 8 |
| Wine | 0.6 | 26 |
| Breast_cancer | 0.5 | 60 |
| Credit_card | 0.34 | 46 |
| Letter | 0.22 | 32 |
| Generated | 0.02 | 4 |

During the execution, we save every clustering result in json files for later analysis. We also save performance and energy results as complementary results in csv files.

We deactivated the use of BLAS backend by setting the OPENBLAS_NUM_THREADS system variable to 1. This library is supposed to enhance scalar calculation like dot products. In our case, it raises errors when using big datasets due to incorrect configuration.

### 5.3 Results

*5.3.1 Repeatability results*

We ran each algorithm on all datasets 30 times and saved every result on a JSON file. For sklearn_dbscan, sklearn_ward and scipy_kmeans, everything is bitwise repeatable. However, we noticed some non-repeatable results in sklearn_kmeans. You can see more details in Table 4, each non-repeatable result is marked with a cross. Tests are achieved with up to 192 threads.

Table 4: K-Means results repeatability for each dataset depending on the number of OpenMP threads used over 30 replications. C=final centers, L=final labels, I=inertia score, M=best iteration. A cross means we found different results in at least two runs out of our 30 replications.

| datasets | nb threads | 1 | | | | 4 | | | | 16 | | | | 64 | | | | 128 | | | | 192 | | | |
|---|---|---|---|---|---|---|---|---|---|---|---|---|---|---|---|---|---|---|---|---|---|---|---|---|---|
| | Results | C | L | I | M | C | L | I | M | C | L | I | M | C | L | I | M | C | L | I | M | C | L | I | M |
| iris | | | | | | | | X | | | | X | | | | X | | | | X | | | | X | |
| wine | | | | | | | | X | | | | X | | | | X | | | | X | | | | X | |
| cancer | | | | | | X | | X | | X | X | X | | X | X | X | | X | X | X | | X | X | X | |
| letter | | | | | | X | | X | | X | X | X | | X | X | X | | X | X | X | | X | X | X | |
| credit_card | | | | | | | | X | | X | X | X | | X | X | X | | X | X | X | | X | X | X | |
| generated | | | | | | X | X | X | | X | X | X | | X | X | X | | X | X | X | | X | X | X | |

When only one thread is used, every result is bitwise repeatable. After four threads, the inertia score is never repeatable across our 30 replications. The inertia score is the sum of squared distances of samples to their closest cluster center. A more paradoxal fact is that sometimes, the final positions of cluster centers is repeatable (for example with the iris and wine datasets) but the inertia score changes! Finally, we notice the resulting centers for the credit_card dataset with 4 threads is repeatable. It seems to be an outlier if we look at the overall tendency.

If we look at the ARI score in Table 5, both K-Means implementations have close results, except for our generated dataset where sklearn_kmeans stands out. We will give an explanation about this in section 6.

Table 5: ARI scores of the tested algorithms for each dataset. As label results are repeatable, we took the first execution of the algorithm applied to the dataset to compute the score

| Datasets \ Algorithms | Scipy_kmeans | Sklearn_kmeans | Sklearn_dbscan | Sklearn_ward |
|---|---|---|---|---|
| **Iris** | 0.7163 | 0.7163 | 0.5464 | **0.7196** |
| **Wine** | 0.8471 | 0.8537 | 0.4259 | **0.9310** |
| **Breast cancer** | **0.7423** | 0.7302 | 0.4877 | 0.5383 |
| **Letter recognition** | 0.1325 | 0.1261 | 0.0317 | **0.1472** |

| Datasets \ Algorithms | Scipy_kmeans | Sklearn_kmeans | Sklearn_dbscan | Sklearn_ward |
|---|---|---|---|---|
| **Default of credit card clients** | 0.0087 | 0.0087 | **0.0140** | 0.0095 |
| **Blob dataset (generated)** | 0.6569 | **0.8828** | 0.7364 | 0.8475 |

*5.3.2 Performance results*

Performance results are given by Figure 3 for the duration of each algorithm over 30 replications and Figure 4 gives the energy consumption of these algorithms over 30 replications.

For the duration, we notice an increase in sklearn_dbscan when we increase the number of threads for iris, wine and breast_cancer datasets. In the worst case, we have for the breast_cancer dataset an increase in time ranging from less than 0.1s with one thread to 1s with more than 128 threads. For letter_recognition, credit_card and generated datasets, the duration decreases when we increase the number of threads up to 16, then it starts to increase. The letter_recognition dataset presents a good performance gain, from 16s with one thread, to less than 2s with 16 threads. The kmeans_scipy algorithm does not seem to have a special behavior with parallelization. Duration results are constant. Same remark for sklearn_ward, results are constant except for breast_cancer dataset with 16 threads. For sklearn_kmeans, increasing the number of threads generally degrades performances. The only exception is letter_recognition dataset for 4 and 16 threads that gives a x2 speed-up. The performance degradation is quite high, with for example a x13 "speed-down" for letter_recognition dataset from 1 to 128 threads meaning too much overhead to handle many threads for this dataset.

For the energy consumption, the overall tendency follows the duration tendency. However, we notice a huge increase in DRAM energy consumption compared to Package energy consumption for the generated dataset with sklearn_ward.

# 6 Discussion

We checked the source code of sklearn_kmeans and found an internal check for parallelization. If the number of data points is not high enough, the number of threads used is downgraded compared to what the user asked for. For example, the iris and wine dataset does not contain enough data point to be parallelized. That explains the repeatable results of the final positions of the centers for these two datasets. However, it seems that the number of threads given by the user is used correctly to compute the inertia, leading to non-repeatable results even for small datasets. Differences in results are at the order of $10^{-14}$. As said at the beginning of this paper, we think this small difference comes from uncontrolled operation order, leading to rounding errors during reduction operations.

When we look to K-Means performances, scipy_kmeans is always faster than sklearn_kmeans, even if the first does not use parallelization. However, this method seems embarrassingly parallel with the use of different initializations. It should definitely take advantage of parallelization. Works on parallelization with OpenMP exist for K-Means [25,26]. They mainly focus on accelerating the assignment or update step by using multiple threads. Datasets used in our experiments had no gain from parallelism. They may be not adapted to the kind of parallelism used by scikit-learn. Regarding the ARI scores, both methods gives similar scores (Table 5) except for the generated dataset. You can see in Figure 2 the original data, and clustering results of both K-Means, alongside the final position of the centers. As K-Means in very sensitive to initialization, we can see that Scipy has fallen into a local minimum compared to scikit-learn which has found a result close to the best solution. Five initializations for Scipy was not enough to achieve the same accuracy as scikit-learn.

The DRAM energy consumption increase for the Ward method on our generated dataset experiment may be due to the counter reset mentioned in section 5.2.1. According to [Intel documentation](), RAPL's package counter "has a wraparound time of around 60 secs when power consumption is high […]". From our tests, we know the reset happens approximatively every 4 000s with our idle system. According to Figure 3, the experiment took 2 400 seconds at each loop. We can think of a case scenario where package energy counters reset twice in an experiment. In this specific case, we would have to add the maximum value a counter is able to store: 262 143 328 850 uJ for each packages results. If we add this value for both package energy consumption result, we obtain a bar chart close to what we already have, with a ratio dram/package following the same tendency as the other experiments on other datasets.

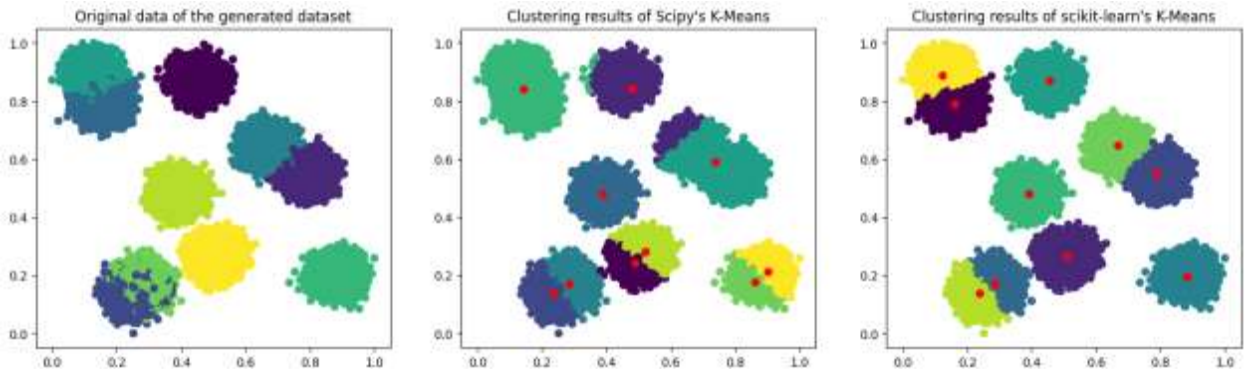

Figure 2: Clustering results of Scipy and scikit-learn's K-Means. Centers results are represented with red dots.

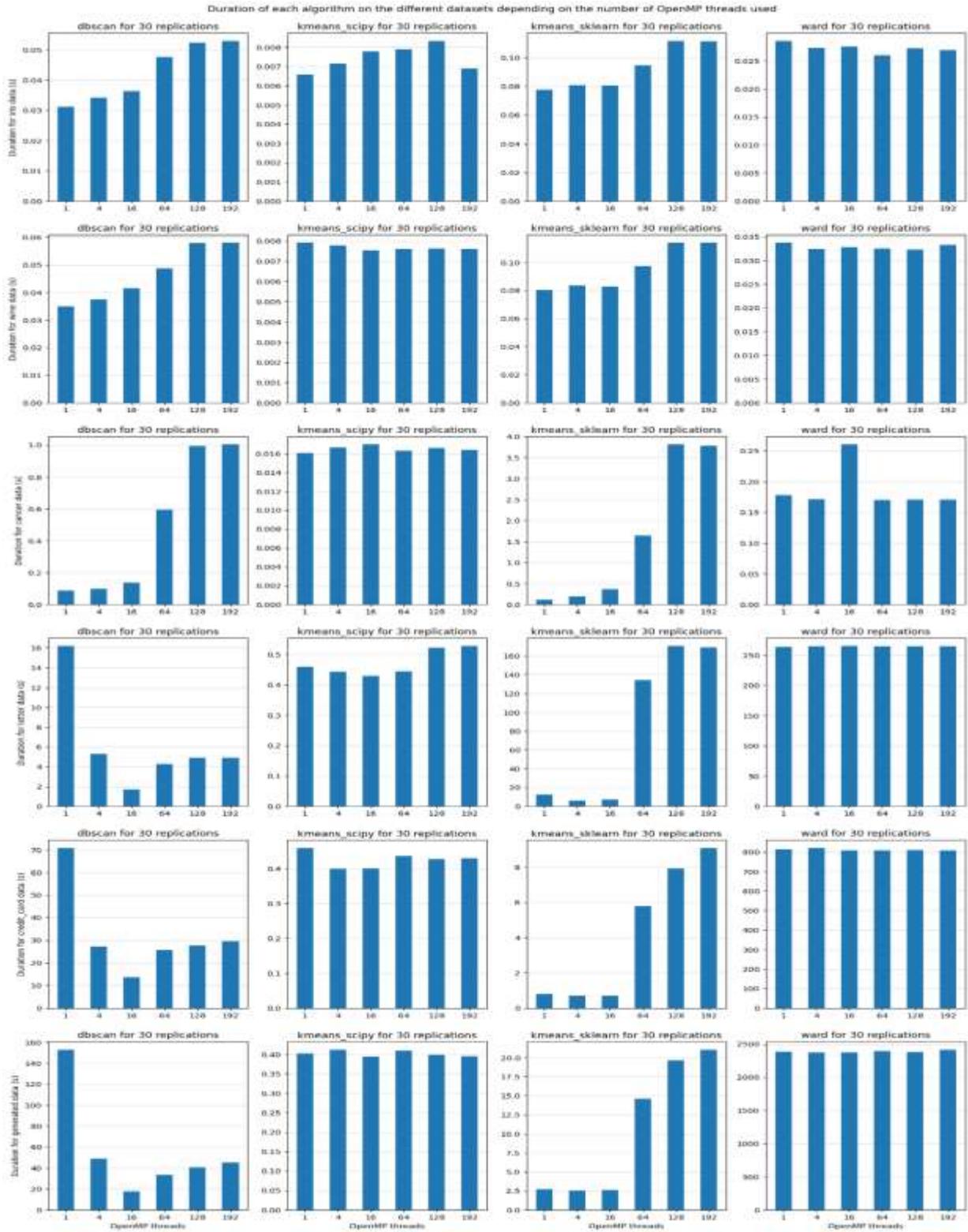

Figure 3: Duration of each algorithm on each dataset depending on the number of OpenMP threads used

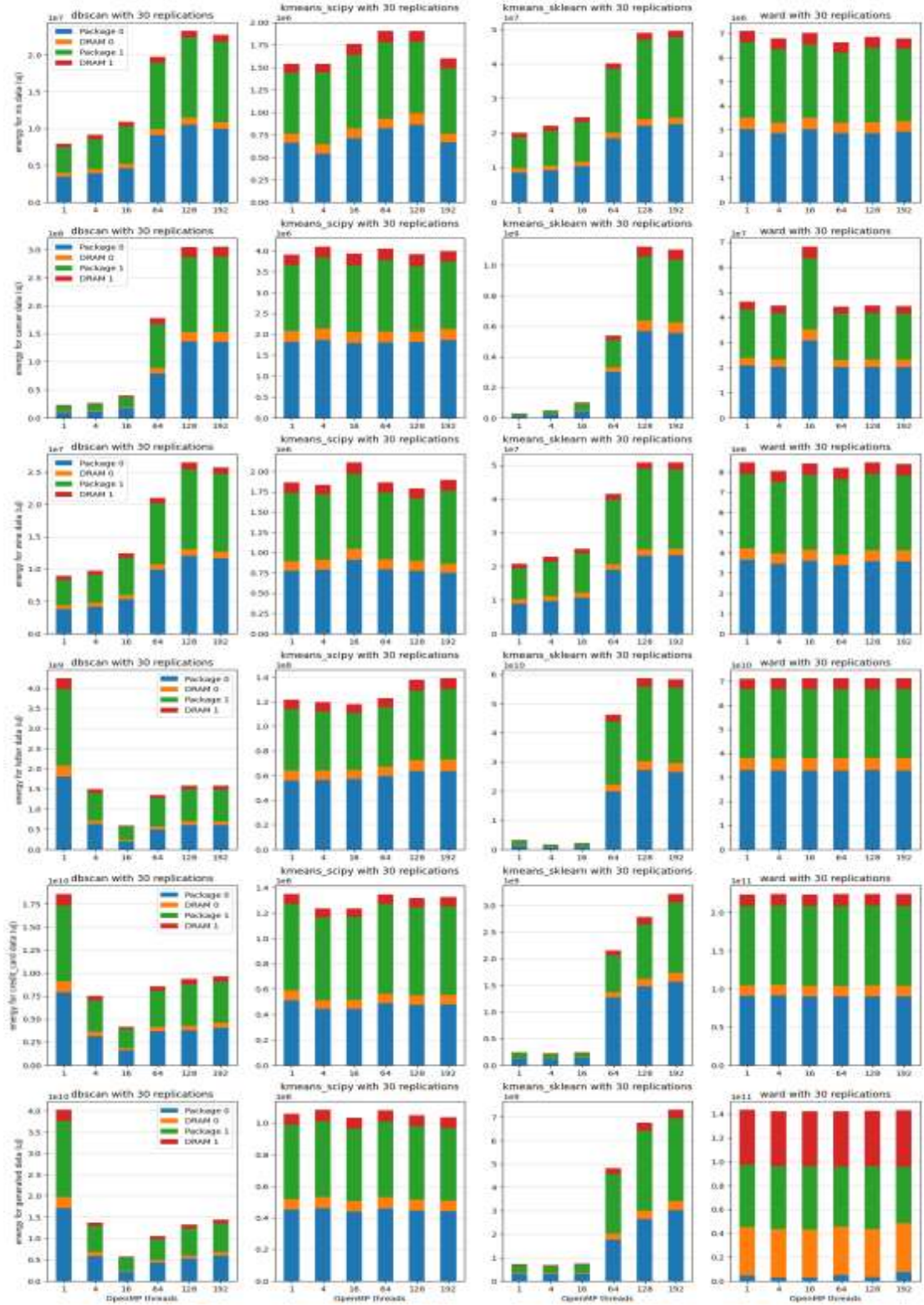

Figure 4: Energy consumption of each algorithm on each dataset depending on the number of OpenMP threads used

# 7 Conclusion

In this paper, we have explored repeatability issues observed in popular clustering methods such as K-means, Ward, and DBSCAN. We identified two repeatability pitfalls that developers should address in order to ensure repeatable behavior at each step of the three aforementioned algorithms. Bitwise repeatability is required primarily for debugging purposes. We found that the scikit-learn implementation of K-means produces non-repeatable results when parallelized using OpenMP. These problems are most likely due to the mismanagement of the OpenMP library, which leads to a random ordering of floating-point operations during computation. We compared this implementation with SciPy's K-means in terms of accuracy, execution time, and energy consumption. The latter is fully repeatable but does not provide a parallel implementation in the tested version. SciPy outperforms scikit-learn in terms of execution time on all tested datasets, while maintaining a comparable ARI score, except for our generated dataset, where it converges to a local minimum compared to scikit-learn using the same hyperparameters.

We believe that many reproducibility problems encountered when using clustering methods are, to a large extent, caused by repeatability issues. These issues may result from a lack of in-depth understanding of the tools, which are often used as black boxes by AI scientists. While practitioners generally have a strong theoretical background in knowledge representation and modeling, and often sufficient programming skills to implement solutions, they do not necessarily possess detailed knowledge of how the underlying machine implementation works, nor of how to write code in a repeatable manner. Such a lack of repeatability can ultimately diminish their scientific contributions. The study presented in [7] reported reproducibility problems mainly due to insufficient documentation and the absence of source code availability (only 6% of the studies provided access to their source code). For studies that did provide source code, it remains unclear what proportion suffers from repeatability issues, as opposed to genuine reproducibility problems such as portability. We aim to increase the visibility of these issues. As future work, we plan to provide a Docker-based version of this experiment to enhance reproducibility. We also intend to improve our measurement tool in order to address the limitations discussed in this paper.

[17] Lloyd, S. (1982). Least squares quantization in PCM. *IEEE Transactions on Information Theory*, *28*(2), 129–137.

[18] Aloise, D., Hansen, P., & Liberti, L. (2012). An improved column generation algorithm for minimum sum-of-squares clustering. *Mathematical Programming*, *131*(1), 195–220.

[19] Alguwaizani, A. (2012). Degeneracy on K-means clustering. *Electronic Notes in Discrete Mathematics*, *39*, 13–20.

[20] Ester, M., Kriegel, H.-P., Sander, J., & Xu, X. (1996). A density-based algorithm for discovering clusters in large spatial databases with noise. *Proceedings of the Second International Conference on Knowledge Discovery and Data Mining*, 226–231.

[21] Beer, A., Draganov, A., Hohma, E., Jahn, P., Frey, C. M. M., & Assent, I. (2023). Connecting the Dots—Density-Connectivity Distance unifies DBSCAN, k-Center and Spectral Clustering. *Proceedings of the 29th ACM SIGKDD Conference on Knowledge Discovery and Data Mining*, 80–92.

[22] Schubert, E., Sander, J., Ester, M., Kriegel, H. P., & Xu, X. (2017). DBSCAN Revisited, Revisited: Why and How You Should (Still) Use DBSCAN. *ACM Trans. Database Syst.*, *42*(3), 19:1-19:21.

[23] Ward Jr., J. H. (1963). Hierarchical Grouping to Optimize an Objective Function. *Journal of the American Statistical Association*, *58*(301), 236–244.

[24] Sander, J., Ester, M., Kriegel, H.-P., & Xu, X. (1998). Density-Based Clustering in Spatial Databases: The Algorithm GDBSCAN and Its Applications. *Data Mining and Knowledge Discovery*, *2*(2), 169–194.

[25] Abdullah, A., H, Q. M., & Ansari, Z. (2021). An OpenMP Based Approach for Parallelization and Performance Evaluation of k-Means Algorithm. *Turkish Journal of Computer and Mathematics Education (TURCOMAT)*, *12*(10), 1524–1537.

[26] Naik, D. S. B., Kumar, S. D., & Ramakrishna, S. V. (2013). Parallel processing of enhanced K-means using OpenMP. *2013 IEEE International Conference on Computational Intelligence and Computing Research*, 1–4.**BIOGRAPHIES**

**VIOLAINE ANTOINE** is an associate professor at Clermont Auvergne University, in the LIMOS laboratory (UMR CNRS 6158). She earned her PdD in 2011 in the University of Compiègne and her Research Director Habilitation in 2023 in the Clermont Auvergne University. Her research interest is focused on data mining and machine learning.

**ANTHONY BERTRAND** is a PhD Student at Clermont Auvergne University (UCA) in the LIMOS laboratory (UMR CNRS 6158). He holds a Master in Computer Science (second head of the list). His thesis subject is about the software-based measurement of energy consumption of Machine Learning programs in High Performance Computing (HPC). He also addresses the reproducibility challenges in the domain of Machine Learning. His email address is anthony.bertrand@uca.fr.

**DAVID R. C. HILL** is a full professor of Computer Science at University Clermont Auvergne (UCA) doing his research at the French Centre for National Research (CNRS) in the LIMOS laboratory (UMR 6158). He earned his Ph.D. in 1993 and Research Director Habilitation in 2000 both from Blaise Pascal University and later became Vice President of this University (2008-2012). He is also past director of a French Regional Computing Center (CRRI) (2008-2010) and was appointed two times deputy director of the ISIMA Engineering Institute of Computer Science – part of Clermont Auvergne INP, #1 Technology Hub in Central France (2005-2007 ; 2018-2021). He is now Director of an international graduate track at Clermont Auvergne INP. Prof Hill has authored or co-authored more than 280 papers and has also published several scientific books. He recently supervised research at CERN in High Performance Computing (David.Hill@uca.fr).

**ENGELBERT MEPHU NGUIFO** is a full professor of computer science at University Clermont Auvergne (UCA), France, where he is the director of Master Degree Program in Computer Science. He is leading research on machine learning and data mining for complex data in the joined University-CNRS laboratory LIMOS where he is co-chair of the Information and Communication Systems research group. His research interests also include formal concept analysis, artificial intelligence, pattern recognition, bioinformatics, big data, and knowledge representation. He was Board member of the French Association on Artificial Intelligence. He is member of the executive board of the French CNRS research group on Artificial Intelligence (GDR RADIA).